\documentclass[letterpaper]{article} 
\usepackage{aaai23}  
\usepackage{times}  
\usepackage{helvet}  
\usepackage{courier}  
\usepackage[hyphens]{url}  
\usepackage{graphicx} 
\usepackage{natbib}  
\usepackage{caption} 
\usepackage{algorithm}
\usepackage{algorithmic}
\usepackage{fancyvrb}
\usepackage{mdframed}
\usepackage{tabularx}
\usepackage{multirow}

\usepackage{rotating} 
\usepackage{cprotect}  

\usepackage{newfloat}
\usepackage{listings}
\DeclareCaptionStyle{ruled}{labelfont=normalfont,labelsep=colon,strut=off} 
\floatstyle{ruled}
\newfloat{listing}{tb}{lst}{}
\floatname{listing}{Listing}
\usepackage{subcaption} 
\newcolumntype{s}{>{\hsize=0.8\hsize}X}
\newcolumntype{B}{>{\hsize=1.2\hsize}X}
\newcommand{\heading}[1]{\multicolumn{1}{c}{#1}}
\title{Forgetful Large Language Models: \\Lessons Learned from Using LLMs in Robot Programming}
\author {
    Juo-Tung Chen,\textsuperscript{\rm 1}
    Chien-Ming Huang \textsuperscript{\rm 1}
}
\affiliations {
    \textsuperscript{\rm 1} Johns Hopkins University, Baltimore, MD 21218, USA\\
    jchen396@jhu.edu, chienming.huang@jhu.edu
}

\begin{document}


\maketitle

\begin{abstract}
Large language models offer new ways of empowering people to program robot applications---namely, code generation via prompting.
However, the code generated by LLMs is susceptible to errors.
This work reports a preliminary exploration that empirically characterizes common errors produced by LLMs in robot programming.
We categorize these errors into two phases: \textit{interpretation} and \textit{execution.} In this work, we focus on errors in execution and observe that they are caused by LLMs being ``forgetful'' of key information provided in user prompts.
Based on this observation, we propose prompt engineering tactics designed to reduce errors in execution.
We then demonstrate the effectiveness of these tactics with three language models: ChatGPT, Bard, and LLaMA-2.
Finally, we discuss lessons learned from using LLMs in robot programming and call for the benchmarking of LLM-powered end-user development of robot applications.






\end{abstract}

\section{Introduction}
Programmable robots have enabled a wide range of applications, ranging from flexible automation to people-facing services. However, programming robot applications effectively requires years of training and experience. The paradigm of end-user programming lowers the barriers to robot programming \cite{c3:ajaykumar2021survey} and empowers end users to develop custom robot applications without substantial engineering training. The rise of large language models introduces new opportunities in this paradigm by offering a natural interface in which end users may program robots \cite{c1:vemprala2023chatgpt}.

However, LLM-powered code generation is not error-free due to its nondeterministic nature \cite{c9:ouyang2023llm}. Despite extensive research efforts aimed at assessing the effectiveness and accuracy of LLM-based code generation tools, certain limitations persist. For instance, these tools may produce inconsistent and occasionally incorrect code outputs. Existing studies have employed approaches such as benchmark evaluations \cite{c6:liu2023your, c7:chen2021evaluating, c8:hammond2021can} and systematic empirical assessments \cite{c10:liu2023no} to explore the capabilities of and challenges in LLM-powered code generation. While these investigations have illuminated various errors and obstacles that may arise during the code generation process, they often fell short in providing comprehensive solutions to enhance code generation stability and minimize the occurrence of errors.

It is worth noting that existing research often focuses primarily on general benchmarking errors, aiming to identify common pitfalls and shortcomings in LLM-generated code; therefore, these studies may not fully capture the specific nuances and intricacies of code specific to a specialized domain such as robotics. As a result, while such benchmark evaluations provide valuable insights into the overall performance of LLMs, they may not comprehensively address the unique challenges posed by code generation for robotic applications.

\begin{figure*}[t]
\centering
\captionsetup{justification=centering} 
    \includegraphics[width=0.8\textwidth]{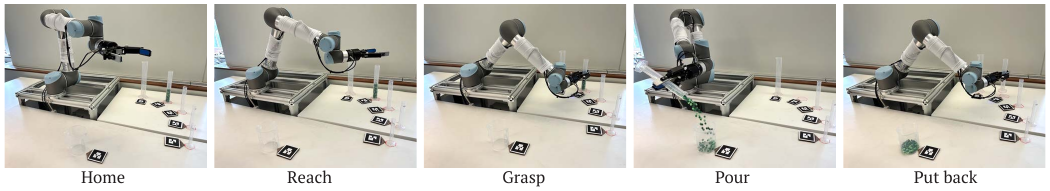}
    \caption{Sequential task execution by the robotic system. The five stages encompass homing, reaching the cylinder, grasping it, pouring its contents into a beaker, and returning the cylinder to its initial position.}
    \label{fig:task-sequence}
\end{figure*}

As a step toward developing the empirical science of incorporating LLMs into robot programming processes, in this work, we sought to explore two research questions:
\textit{1) What are the common errors produced by LLMs in end-user robot programming?} and
\textit{2) What practical strategies can be employed to mitigate and reduce these errors?}
To ground our exploration, we designed a sequential manipulation task (Figure \ref{fig:task-sequence}) and tested three language models---ChatGPT, Bard, and LLaMA-2---to assess their capabilities in generating code to complete the task.

Our key findings are 
1) LLMs are ``forgetful'' and do not consider information provided in the system prompt as hard fact;
2) the forgetfulness of LLMs leads to errors in code execution;
3) in addition to execution errors, LLMs make various errors (e.g., syntax errors, missing necessary libraries) that cause failures in code interpretation; and
4) simple strategies---such as reinforcing task constraints in the objective prompt and extracting numerical task contexts from the system prompt and storing them in data structures---seem to notably reduce execution errors caused by LLM forgetfulness.

\section{Experiment 1: Identifying Common Errors}

\subsection{Programming Task}
In order to assess the code generation ability and performance of LLMs in robot programming, we set up a sequential manipulation task.
Our experimental setup includes a UR5 manipulator paired with a webcam for basic perception via AR markers, allowing for the registration of task objects into a virtual workspace for precise and accurate motion planning via MoveIt. 
The sequential manipulation task involves the robot picking up a graduated cylinder and pouring its contents into a beaker; this task is a common step in biochemical lab tests\footnote{We envision the automation of several biochemical lab tests through custom robot applications so as to accelerate scientific experimentation.}. 
The high-level procedure of the manipulation task involves: 
\begin{enumerate}
\item Moving the robot to a home (neutral) position;
\item Reaching out to the graduated cylinder;
\item Grasping the graduated cylinder at its midpoint;
\item Performing the pouring action (including moving to the target location and rotating the robot's end effector); and
\item Placing the cylinder back in its original position.
\end{enumerate}


\subsection{Baseline Prompt}
A descriptive prompt is believed to enhance the quality of LLM-generated responses. 
It has been documented that a well-constructed prompt should contain the following components \cite{c1:vemprala2023chatgpt}: constraints and requirements, environmental description, current state of the system, goals and objectives, description of the robotic API library, and solution examples. 
Consequently, our baseline prompt is composed of four parts: \textbf{system prompt, description of robotic API library, solution example,} and \textbf{objective prompt.} See the appendix for the full baseline prompt used in our experiments.

\subsubsection{System Prompt}
Here, we defined the role of the LLM and provided it with task constraints and requirements. We additionally included contextual details regarding the environment to alert the LLM to potential task objects. 

\subsubsection{Description of Robotic API Library}
We provided a clear rundown of how each high-level function provided for the LLM should be used, along with useful reminders and conventions. It is worth noting that by providing descriptive names for all of the API functions, the LLM's ability to understand the functional links between APIs may be enhanced, which can facilitate the LLM to produce more desirable outcomes for the given problem \cite{c1:vemprala2023chatgpt}.

\subsubsection{Solution Example}
We provided an example solution to guide the LLM's solution strategy and to (hopefully) prevent it from generating erroneous responses.

\subsubsection{Objective Prompt}
Here, we articulated the intended objective for the LLM to respond to while considering all prompts as outlined previously. Below is the objective prompt used in our experiments:

\begin{mdframed}[linewidth=0.3pt, innerleftmargin=8pt, innerrightmargin=8pt, innertopmargin=8pt, innerbottommargin=8pt, roundcorner=5pt, splittopskip=15pt, splitbottomskip=10pt, nobreak=false]
\begin{footnotesize}
Please write a Python function to pick up a 25mL graduated cylinder at Marker 15 and pour its contents into a 500mL beaker at Marker 7. After that, put the cylinder back to where it was.
\end{footnotesize}
\end{mdframed}

\subsection{Large Language Models}
In our experiments, we used three language models: ChatGPT (3.5-turbo-0613), Bard, and LLaMA-2 (13B parameters).
Given the stochastic nature of these LLMs, each model was tested ten times while keeping the prompts and sequential manipulation task the same across trials.

\begin{figure*}[t]
    \centering 

    \includegraphics[width=0.9\textwidth]{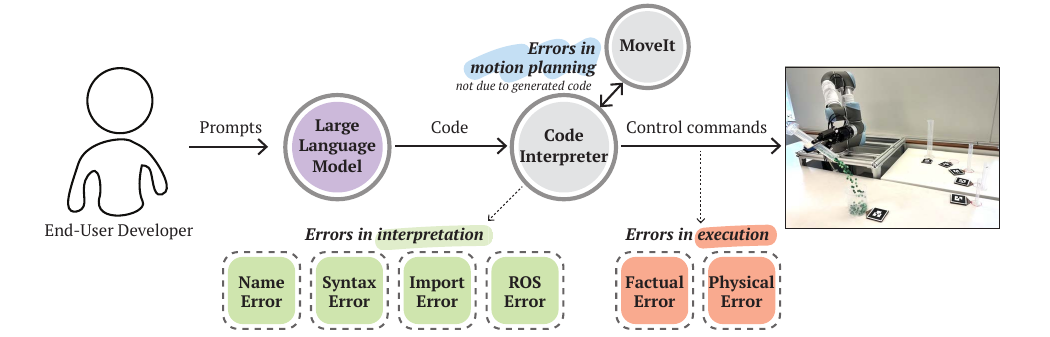}
    \captionsetup{justification=centering} 
    
    \caption{Workflow and the emergence of potential errors in utilizing an LLM in robot programming.}
    \label{fig:error-overview}
\end{figure*}

\subsection{Findings}
Our first experiment sought to understand common errors produced by the three language models. 
To this end, we manually characterized the observed errors, which
can be grouped roughly into two categories representing errors in different phases of application development---\textbf{errors in interpretation} and \textbf{errors in execution}---as illustrated in Figure \ref{fig:error-overview}. We note that there may be errors in motion planning that have nothing to do with LLM-generated code, which is outside the scope of this work.


\subsubsection{Errors in Interpretation}
\begin{figure}[t]
    \centering 
    \includegraphics[width=0.4\textwidth]{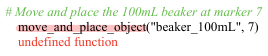}
    \captionsetup{justification=centering} 
    \caption{Name error (using undefined functions).}
    \label{fig:error-name}
\end{figure}

\begin{figure}[t]
    \centering 
    \includegraphics[width=0.35\textwidth]{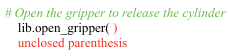}
    \captionsetup{justification=centering} 
    \caption{Syntax error (syntactically incorrect).}
    \label{fig:error-syntax}
\end{figure}

\begin{figure}[t]
    \centering 
    \includegraphics[width=0.4\textwidth]{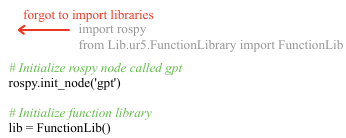}
    \captionsetup{justification=centering} 
    \caption{Import error (oversight in importing necessary libraries).}
    \label{fig:error-import}
\end{figure}

\begin{figure}[t]
    \centering 
    \includegraphics[width=0.4\textwidth]{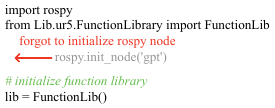}
    \captionsetup{justification=centering} 
    \caption{ROS error (omission of ROS node initialization).}
    \label{fig:error-ros}
\end{figure}

Errors in this category cause failures in code interpretation and include four different subtypes:
    \begin{enumerate}
        \item[\textbf{a.}] \textbf{Name Error:} This error type includes instances where references to variables or functions precede their definition or initialization within the code (Figure \ref{fig:error-name}).
        \item[\textbf{b.}] \textbf{Syntax Error:} Characterized by syntactically incorrect code structures, this error type hinders the proper interpretation of the generated code (Figure \ref{fig:error-syntax}).
        \item[\textbf{c.}] \textbf{Import Error:} This error type typically indicates that the generated code does not include the necessary libraries for code interpretation (Figure \ref{fig:error-import}).
        \item[\textbf{d.}] \textbf{ROS Error:} Within the context of the Robot Operating System, this type of error surfaces due to the omission of ROS node initialization or incorrect utilization of ROS packages, negatively impacting the overall communication and coordination within the robotic system (Figure \ref{fig:error-ros}).
    \end{enumerate}


\begin{figure}[t]
    \centering 
    \includegraphics[width=0.3\textwidth]{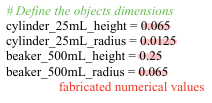}
    \captionsetup{justification=centering} 
    \caption{Factual error (using fabricated numerical values to define the objects' dimensions).}
    \label{fig:error-factual}
\end{figure}

\begin{figure}[t]
    \centering 
    \includegraphics[width=0.4\textwidth]{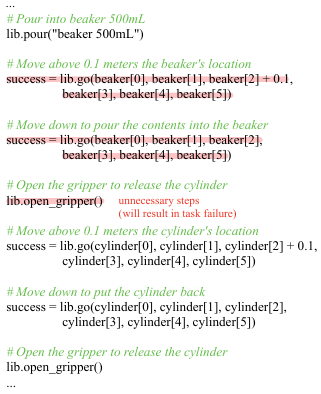}
    \captionsetup{justification=centering} 
    \caption{Physical error (impractical physical action for the given task).}
    \label{fig:error-physical}
\end{figure}

    
\subsubsection{Errors in Execution} Errors in this category cause failures in code execution---even though the code may be interpretable---and include two types:  
    \begin{enumerate}
        \item[a.] \textbf{Factual Error:} This error type indicates model hallucination; for example, instead of using numerical values that the user provides in the system prompt to describe the task objects, the model fabricates numbers, subsequently causing errors in motion planning or execution (Figure \ref{fig:error-factual}).  
        \item[b.] \textbf{Physical Error:} This error type includes errors that ultimately cause execution failures even if all other error types are not present. Examples include adding unnecessary steps to the action sequence (Figure \ref{fig:error-physical}).
    \end{enumerate}

The two error categories---\textit{interpretation} and \textit{execution}---call for different methods of error handling. Errors in interpretation are typically caught by the program interpreter or compiler, which displays error messages that help users address the errors in a more straightforward identification and rectification process \cite{c2:inagaki2023llms}. 
In contrast, errors in execution are less obvious, as they do not necessarily cause immediate code breakdown; these errors surface only when undesirable task outcomes are observed.


\begin{table*}
\small
\centering
\renewcommand{\arraystretch}{1.0}
\begin{tabularx}{\linewidth}{c B c s c s c X}
\hline
Model & \heading{ChatGPT} & & \heading{Bard} & & \heading{LLaMA-2} & \\
\cline{2-3} \cline{4-5} \cline{6-7}
Trial & Types of error & Completion & Types of error & Completion & Types of error & Completion \\
\hline
1 & Factual, Physical & No & Factual, Physical & No & Factual, Physical & No \\
2 & Factual, Physical & No & Factual, Physical & No & Factual, Physical & No \\
3 & Factual, Physical & No & Factual, Physical & No & Factual, Physical & No \\
4 & Factual, Physical & No & Factual, Physical & No & Factual, Physical & No \\
5 & Factual, Physical & No & Factual, Physical & No & Factual, Physical & No \\
6 & Factual, Physical, Import, ROS & No & Factual, Physical & No & Factual, Physical & No \\
7 & Factual, Physical & No & Factual, Physical & No & Physical, Name & No \\
8 & Factual, Physical, Import, ROS & No & Factual, Physical & No & Factual & No \\
9 & Factual, Import, ROS & No & Factual, Physical & No & Factual, Physical & No \\
10 & Factual, Physical, Import, ROS & No & Factual, Physical & No & Factual, Physical & No \\
\hline
\end{tabularx}
\caption{Common error identification experiment results.}
\label{tab:baseline-results}
\end{table*}

Our experiment revealed varying patterns of error occurrence across the three language models (Table \ref{tab:baseline-results}). 
To our surprise, none of the three models successfully completed the intended task in any of the trials. This result underscores the challenges involved in translating end-user prompts into accurate and executable robot control code via LLMs. 
Furthermore, across the three models evaluated, \textit{factual} and \textit{physical} errors were most common; the prevalence of these errors highlights a key limitation of LLM-based code generation for end-user development of robot applications, which prompted us to explore practical strategies to reduce these types of execution errors.

\begin{figure}[t]
\centering
    \includegraphics[width=0.45\textwidth]{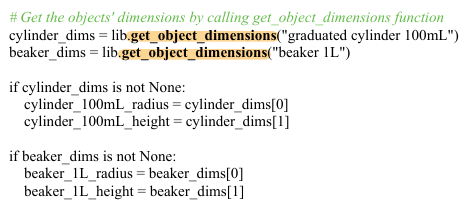}    
    \caption{In the robotic API library, we provided a dedicated function for parsing the system prompt and retrieving the objects' dimensions. Its usage was provided correspondingly in the solution example prompt.}
    \label{fig:strategy-1}
\end{figure}

\begin{figure}[t]
\centering
    \includegraphics[width=0.45\textwidth]{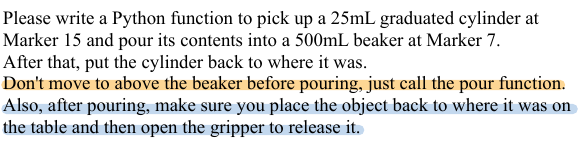}    
    \caption{In the objective prompt, we added a sentence to reinforce constraints (orange) and another sentence to articulate the physical implications (blue).}
    \label{fig:strategy-2}
\end{figure}

\section{Experiment 2: Exploring Practical Strategies to Reduce Errors in Execution}
This experiment studied strategies that might enhance an LLM's ability to generate accurate and reliable code for robotic applications.
This experiment followed the same protocol (e.g., same manipulation task, ten trials per language model) as the first experiment.


\subsection{Practical Strategies}
In Experiment 1, we found that errors in execution may be attributed to the ``forgetfulness'' of LLMs; the models appear to ``forget'' the information provided in user prompts or do not treat the provided description as factual information to use in code generation. 
Therefore, we explored the following strategies' effectiveness in addressing the issue of forgetfulness: 

\begin{enumerate}
    \item When prompts involve task/context information specified in numerical form, implement dedicated functions for retrieving precise, numerical data. (Figure \ref{fig:strategy-1})
    \item When dealing with intricate functions (like the pour function in our experiment), reinforce key constraints in the objective prompt to ensure more accurate and reliable code generation. (Figure \ref{fig:strategy-2})
\end{enumerate}

In addition to the these strategies, enhancing the clarity and specificity of the objective prompt by articulating its physical implications or providing greater descriptive context can also help curtail excessive divergence in LLM-generated code. 
Our implementations of these strategies are shown in Figures \ref{fig:strategy-1} and \ref{fig:strategy-2}.

\subsection{Findings}
Table \ref{tab:strategic-results} shows the results of adopting the strategies proposed above. 
Across all models, we observed a substantial increase in successful task completion and a decrease in the number of \textit{factual} and \textit{physical} errors.
Specifically, ChatGPT was able to achieve a task completion rate of 60\% and 
errors in execution were reduced by 94.7\% as compared to its results in Experiment 1. 
Bard achieved a similar success rate of 70\% with strategy implementation and the occurrence of factual and physical errors was reduced by 95\%.
However, LLaMA-2-13B only reached a task completion rate of 40\% using the strategies and factual and physical errors were reduced by only 83.3\%.



\begin{table*}
\centering
\small 
\renewcommand{\arraystretch}{1.0} 
\begin{tabularx}{\linewidth}{c X c s c s c X}
\hline
Model & \heading{GPT 3.5} & & \heading{Bard} & & \heading{LLaMA} & \\
\cline{2-3} \cline{4-5} \cline{6-7}
Trial & Types of error & Completion & Types of error & Completion & Types of error & Completion \\
\hline
1 & None & \textbf{Yes} & None & \textbf{Yes} & None & \textbf{Yes} \\ 
2 & None & \textbf{Yes} & None & \textbf{Yes} & Name & No \\ 
3 & Import, ROS, Factual & No & Import, ROS, Factual & No & None & \textbf{Yes} \\ 
4 & Name & No & Name & No & None & \textbf{Yes} \\ 
5 & Import, ROS, Name & No & None & \textbf{Yes} & Name, Physical & No \\ 
6 & None & \textbf{Yes} & None & \textbf{Yes} & Name, Physical & No \\ 
7 & None & \textbf{Yes} & None & \textbf{Yes} & Name, Physical & No \\ 
8 & Name & No & None & \textbf{Yes} & None & \textbf{Yes} \\ 
9 & None & \textbf{Yes} & None & \textbf{Yes} & Name & No \\ 
10 & None & \textbf{Yes} & None & \textbf{Yes} & Name & No \\ 
\hline
\end{tabularx}
\caption{Strategic Prompting Experiment Results}
\label{tab:strategic-results}
\end{table*}




\section{Discussion}

\subsection{Lessons Learned}
While promising, LLM-based code generation for end-user development of robot applications remains inconsistent, which is unsurprising given the intricate and probabilistic design of these models. This work highlights the importance of keeping users in the loop in application development.

We additionally determined that the success of LLM-powered code generation often hinges on the user's ability to provide explicit and descriptive objective prompts; for instance, specifying detailed instructions such as ``Place the cylinder back to its original position'' yields more accurate results than ambiguous directives like ``Put it back.''

Furthermore, we found that errors in execution primarily stem from the forgetfulness of LLMs, which causes them to overlook information supplied in prompts. 
Consequently, we made a concerted effort to explicitly emphasize the instruction, ``All the information I provided should be treated as factual information and shouldn't be ignored.'' Despite this explicit instruction, unsatisfactory outcomes persisted, indicating that simple reinforcement is ineffective. 

Lastly, a suite of tools is needed for the productive use of LLM-based robot programming: at the basic level, custom verification scripts may be used to identify and correct errors in interpretation (e.g., missing libraries); the strategies discussed in this work may also help reduce factual and physical errors; and a preview tool may allow users to simulate program behavior prior to robot deployment, thereby reducing unforeseen errors during actual execution. 


\subsection{Call for Benchmarks}
In light of the evolving landscape of LLM-driven robot programming, we advocate for the establishment of standardized benchmarks that encompass a diverse set of tasks and metrics to assess the performance of LLMs in various programming scenarios. Such benchmarks will let researchers, practitioners, and developers collectively advance the science of LLM-driven robot programming.

\subsection{Limitations and Future Work}
This preliminary work has limitations that may motivate future research.
Our experiments focused on a single manipulation task, which does not capture the vast array of scenarios in end-user robot programming. Future work may build on our exploration and include a wider range of representative programming tasks and language models.

In our experiments, we simplified the challenges of robot perception by using AR markers. As new vision-language models are developed, future research should study the true complexity of incorporating large data models in the various processes of robot programming.

Future work should also include a comprehensive evaluation of different aspects of end-user robot programming, including debugging; we speculate that debugging may be particularly challenging in the new paradigm of LLM-powered robot programming, as end users will need to spend time understanding the generated code and developing a mental model of it in order to resolve errors successfully.

    


\section*{Acknowledgments}
This work was supported by National Science Foundation award \#2143704. The authors would like to thank Jaimie Patterson and Ulas Berk Karli for their help with this work.


\begin{thebibliography}{8}
\providecommand{\natexlab}[1]{#1}

\bibitem[{Ajaykumar, Steele, and Huang(2021)}]{c3:ajaykumar2021survey}
Ajaykumar, G.; Steele, M.; and Huang, C.-M. 2021.
\newblock A survey on end-user robot programming.
\newblock \emph{ACM Computing Surveys (CSUR)}, 54(8): 1--36.

\bibitem[{Chen et~al.(2021)Chen, Tworek, Jun, Yuan, Pinto, Kaplan, Edwards,
  Burda, Joseph, Brockman et~al.}]{c7:chen2021evaluating}
Chen, M.; Tworek, J.; Jun, H.; Yuan, Q.; Pinto, H. P. d.~O.; Kaplan, J.;
  Edwards, H.; Burda, Y.; Joseph, N.; Brockman, G.; et~al. 2021.
\newblock Evaluating large language models trained on code.
\newblock \emph{arXiv preprint arXiv:2107.03374}.

\bibitem[{Hammond~Pearce et~al.(2021)Hammond~Pearce, Ahmad, Karri, and
  Dolan-Gavitt}]{c8:hammond2021can}
Hammond~Pearce, B.~T.; Ahmad, B.; Karri, R.; and Dolan-Gavitt, B. 2021.
\newblock Can openai codex and other large language models help us fix security
  bugs.
\newblock \emph{arXiv preprint arXiv:2112.02125}.

\bibitem[{Inagaki et~al.(2023)Inagaki, Kato, Takahashi, Ozaki, and
  Kanda}]{c2:inagaki2023llms}
Inagaki, T.; Kato, A.; Takahashi, K.; Ozaki, H.; and Kanda, G.~N. 2023.
\newblock LLMs can generate robotic scripts from goal-oriented instructions in
  biological laboratory automation.
\newblock \emph{arXiv preprint arXiv:2304.10267}.

\bibitem[{Liu et~al.(2023{\natexlab{a}})Liu, Xia, Wang, and
  Zhang}]{c6:liu2023your}
Liu, J.; Xia, C.~S.; Wang, Y.; and Zhang, L. 2023{\natexlab{a}}.
\newblock Is your code generated by chatgpt really correct? rigorous evaluation
  of large language models for code generation.
\newblock \emph{arXiv preprint arXiv:2305.01210}.

\bibitem[{Liu et~al.(2023{\natexlab{b}})Liu, Tang, Luo, Zhou, and
  Zhang}]{c10:liu2023no}
Liu, Z.; Tang, Y.; Luo, X.; Zhou, Y.; and Zhang, L.~F. 2023{\natexlab{b}}.
\newblock No Need to Lift a Finger Anymore? Assessing the Quality of Code
  Generation by ChatGPT.
\newblock \emph{arXiv preprint arXiv:2308.04838}.

\bibitem[{Ouyang et~al.(2023)Ouyang, Zhang, Harman, and
  Wang}]{c9:ouyang2023llm}
Ouyang, S.; Zhang, J.~M.; Harman, M.; and Wang, M. 2023.
\newblock LLM is Like a Box of Chocolates: the Non-determinism of ChatGPT in
  Code Generation.
\newblock \emph{arXiv preprint arXiv:2308.02828}.

\bibitem[{Vemprala et~al.(2023)Vemprala, Bonatti, Bucker, and
  Kapoor}]{c1:vemprala2023chatgpt}
Vemprala, S.; Bonatti, R.; Bucker, A.; and Kapoor, A. 2023.
\newblock Chatgpt for robotics: Design principles and model abilities.
\newblock \emph{Microsoft Auton. Syst. Robot. Res}, 2: 20.

\end{thebibliography}

\clearpage
\appendix
\section{Appendix: Initial Prompt}
\label{appendix:SystemPrompt}%
\subsection{System Prompt}
\begin{mdframed}[linewidth=0.3pt, innerleftmargin=8pt, innerrightmargin=8pt, innertopmargin=8pt, innerbottommargin=8pt, roundcorner=5pt, splittopskip=15pt, splitbottomskip=10pt, nobreak=false]%
\begin{scriptsize}

You are an assistant helping me with the UR5 robot arm.
This is a 6 degrees of freedom robot manipulator that has a gripper as its end effector.
The gripper is in the open position in the beginning.
When I ask you to do something, you are supposed to give me Python code that is needed to achieve that task using the UR5 robot arm and then an explanation of what that code does.
You are only allowed to use the functions I have defined for you.
You are not to use any other hypothetical functions that you think might exist.
You can use simple Python functions from libraries such as math and numpy.
You should put all the code in one block and put the explanation after the code. Don't break the code into pieces.
Always remember to import the function FunctionLib.
Always use floating numbers; for example, instead of 2, use 2.0. \\
In the environment, the following items might be present: \\
beaker 1L: radius = 6.5 cm and height = 15 cm , \\
beaker 500mL: radius = 5.5 cm and height = 12 cm,\\
beaker 250mL: radius = 4.75 cm and height = 10 cm,\\
beaker 100mL: radius = 3 cm and height = 7 cm,\\
beaker 50mL: radius = 2.5 cm and height = 5.5 cm,\\
graduated cylinder 250mL: radius = 4 cm and height = 31 cm,\\
graduated cylinder 100mL: radius = 3.25 cm and height = 25.5 cm,\\
graduated cylinder 50mL: radius = 3 cm and height = 19 cm,\\
graduated cylinder 25mL: radius = 2.5 cm and height = 15.5 cm,\\
graduated cylinder 10mL: radius = 2 cm and height = 14 cm.\\
Use the dimensions provided above when I don't specifically tell you the dimensions of the object.
\end{scriptsize}
\end{mdframed}

\subsection{Description of Robotic API Library}

\begin{mdframed}[linewidth=0.3pt, innerleftmargin=8pt, innerrightmargin=8pt, innertopmargin=8pt, innerbottommargin=8pt, roundcorner=5pt, splittopskip=15pt, splitbottomskip=10pt, nobreak=false] 
\begin{scriptsize}
At any point, you have access to the following functions, which are accessible after initializing a function library. You are not to use any hypothetical functions. All units are in the SI system.\\
\verb|lib=FunctionLib()|: Initializes all functions; access any of the following functions by using lib.\\
\verb|move_to_home_position()|: Moves the robot to a neutral home position.\\
\verb|get_marker_location(marker_number)|: Given an integer marker ID such as 1, gets the x, y, z coordinates of the marker with respect to the base frame in meters, and roll, pitch, yaw in degrees with respect to the base frame of the robot.\\
\verb|go(x,y,z,roll,pitch,yaw)|: Moves the robot arm to the x, y, z position in meters, and roll, pitch, yaw in degrees with respect to the base frame of the robot.\\
\verb|get_current_end_effector_pose()|: Returns the current end effector pose in x, y, z positions in meters and roll, pitch, yaw in degrees.\\
\verb|open_gripper()|: Opens the gripper.\\
\verb|close_gripper(name)|: Gets a string for the name of the object to be grasped and the width of the object in meters as a floating number. Closes the gripper.\\
\verb|add_cylinder_to_workspace(name,x,y,z,height,radius)|: Adds a cylinder to the virtual workspace; must be done before planning to move. Name is a string and x,y,z are the locations of the object in meters. Height and radius are in meters.\\
\verb|pour(target_container_name)|: The robot will go to near the target container and rotate its wrist to pour the contents inside the object that is grasped by the gripper into the target container.

A few useful things:
Always start your code by importing FunctionLib and also make sure to always init a node with rospy.
If you are uncertain about something, you can ask me a clarification question, as long as you specifically identify it by saying ``Question."
Here is an example scenario that illustrates how you can ask clarification questions. Let us assume a scene contains two beakers of different sizes.

Me: Go and grab the beaker and then come back to this location.

You: Question: There are two beakers. Which one do you want me to go to?

Me: 500mL beaker, please.

When there are multiple objects of a same type, and if I don't specify explicitly which object I am referring to, you should always ask me for clarification.
Never make assumptions.

In terms of axis conventions, forward means positive X-axis. Right means positive Y-axis. Up means positive Z-axis.

Also, while using \verb|get_marker_location|, sometimes the marker might not exist in the scene (occluded); the function will return ``None'' in this case. Therefore, you should check if pose is ``None'' before using it.
This way, you can handle the case where the marker's tf transform is not found and avoid raising an error. You can print a message or perform any other desired actions to handle the absence of the marker pose.
\end{scriptsize}
\end{mdframed}

\subsection{Solution Example}

\begin{mdframed}[linewidth=0.3pt, innerleftmargin=8pt, innerrightmargin=8pt, innertopmargin=8pt, innerbottommargin=8pt, roundcorner=5pt, splittopskip=15pt, splitbottomskip=10pt, nobreak=false]
\begin{scriptsize}
    
The following is an example of writing the code. If the user asked, ``There is a 100mL graduated cylinder on Marker 6 and a 1L beaker on Marker 9. Pick up the graduated cylinder and pour its contents into the beaker. After pouring, place the graduated cylinder at Marker 5," then you should write a code like the following:
\label{lst:listing}%
\begin{lstlisting}[basicstyle={\tiny}]

from Lib.ur5.FunctionLibrary import FunctionLib
import rospy

# Initialize rospy node called gpt
rospy.init_node(`gpt')

# Initialize function library
lib = FunctionLib()

# Move the robot back to home position
lib.move_to_home_position()
rospy.sleep(2)

# Defining the objects' dimensions
cylinder_height = 0.255
cylinder_radius = 0.0325
beaker_height = 0.15
beaker_radius = 0.065

# Get the locations of Marker 6, Marker 9, and Marker 5
marker_6_location = lib.get_marker_location(6)
marker_9_location = lib.get_marker_location(9)
marker_5_location = lib.get_marker_location(5)

# Check if the markers are visible
if marker_6_location is None:
    print(``Marker 6 not found. Please check the environment")
    exit()
if marker_9_location is None:
    print(``Marker 9 not found. Please check the environment")
    exit()
if marker_5_location is None:
    print(``Marker 5 not found. Please check the environment")
    exit()
    
# Add the graduated cylinder 100mL into the workspace
lib.add_cylinder_to_workspace(``grad_cylinder_100mL", marker_6_location[0], marker_6_location[1], marker_6_location[2] + cylinder_height / 2.0, cylinder_height, cylinder_radius)

# Add the beaker 500mL into the workspace
lib.add_cylinder_to_workspace(``beaker_500mL", marker_9_location[0], marker_9_location[1], marker_9_location[2] + beaker_height / 2.0, beaker_height, beaker_radius)

# Get the locations of the objects
cylinder = lib.get_object_location(``grad_cylinder_100mL")
beaker = lib.get_object_location(``beaker_500mL")

# Move above 0.1 meters the cylinder's location
success = lib.go(cylinder[0], cylinder[1], cylinder[2] + 0.1, 
                 cylinder[3], cylinder[4], cylinder[5]) 

# Move down to grasp the cylinder
success = lib.go(cylinder[0], cylinder[1], cylinder[2], 
                 cylinder[3], cylinder[4], cylinder[5])

# Close the gripper to grasp the cylinder
lib.close_gripper(``graduated cylinder 100mL")

# Pour into beaker 500mL
lib.pour(``beaker 500mL")

# Move to Marker 5
success = lib.go(marker_5_location[0], marker_5_location[1], marker_5_location[2] + cylinder_height / 2.0, 
                 marker_5_location[3], marker_5_location[4], marker_5_location[5])

# Open the gripper to release the cylinder
lib.open_gripper()

print(``Task finished")

\end{lstlisting}
This code picks up the 100mL graduated cylinder, pours it into the 1L beaker, and then places it at Marker 5.
\end{scriptsize}

\end{mdframed}

\end{document}